\documentclass[runningheads]{llncs}
\usepackage{graphicx}

\usepackage{tikz}
\usepackage{comment}
\usepackage{amsmath,amssymb} 
\usepackage{color}
\usepackage{dsfont}
\usepackage{soul}

\usepackage[accsupp]{axessibility}  

\usepackage[width=122mm,left=12mm,paperwidth=146mm,height=193mm,top=12mm,paperheight=217mm]{geometry}

\newcommand{\eg}{e.g.,\ }
\newcommand{\ie}{i.e.,\ }
\newcommand{\etal}{et~al.\ }

\begin{document}


\pagestyle{headings}
\mainmatter
\def\ECCVSubNumber{}  




\title{Neural Radiance Transfer Fields for Relightable Novel-view Synthesis with Global Illumination} 
\titlerunning{Neural Radiance Transfer Fields} 

\author{Linjie Lyu$^1$, Ayush Tewari$^2$, Thomas Leimk\"uhler$^1$, Marc Habermann$^1$, and Christian Theobalt$^1$}
\authorrunning{Lyu et al.}

\institute{$^1$Max Planck Institute for Informatics, Saarland Informatics Campus \\ $^2$MIT}

\maketitle

%
%
\begin{figure}
    \vspace{-12pt}
	\begin{center}
		\includegraphics[width=\linewidth]{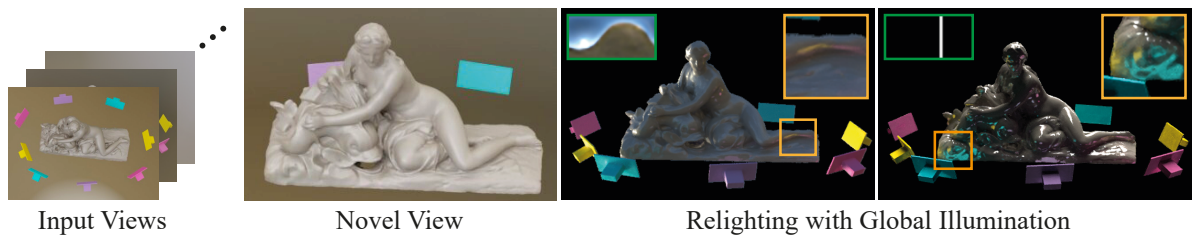}
	\end{center}
	\vspace{-12pt}
	\caption
	{
	Our method takes multiple views of a scene under one unknown illumination condition as input and allows novel-view synthesis and relighting (corresponding environment maps in green insets) with intricate multi-bounce illumination (orange insets).
	}
	\label{fig:teaser}
	\vspace{-24pt}
\end{figure}
%
%


\begin{abstract}
Given a set of images of a scene, the re-rendering of this scene from novel views and lighting conditions is an important and challenging problem in Computer Vision and Graphics. 
On the one hand, most existing works in Computer Vision usually impose many assumptions regarding the image formation process, e.g. direct illumination and predefined materials, to make scene parameter estimation tractable.
On the other hand, mature Computer Graphics tools allow modeling of complex photo-realistic light transport given all the scene parameters.
Combining these approaches, we propose a method for scene relighting under novel views by learning a neural precomputed radiance transfer function, which implicitly handles global illumination effects using novel environment maps.
Our method can be solely supervised on a set of real images of the scene under a single unknown lighting condition.
To disambiguate the task during training, we tightly integrate a differentiable path tracer in the training process and propose a combination of a synthesized OLAT and a real image loss.
Results show that the recovered disentanglement of scene parameters improves significantly over the current state of the art and, thus, also our re-rendering results are more realistic and accurate.
\vspace{-12pt}
\end{abstract}

%
\section{Introduction}
\label{sec:introduction}
%
%
The image formation process is influenced by many factors such as the scene geometry, the object materials, the lighting, and the properties of the recording camera. 
Recovering these properties solely from the final images of the scene is an important inverse problem in Computer Vision and enables several applications such as scene understanding, virtual reality, and controllable image synthesis. 
Since this is an ill-posed and challenging inverse problem, existing methods make several assumptions about the 3D scene. 
Common assumptions are that scenes are diffuse~\cite{wu2020unsupervised}, or can be described by some predefined material models~\cite{zhang2021nerfactor,physg2020}.
Importantly, most methods only consider the direct scene illumination~\cite{philip2019multi,zhang2021nerfactor,Boss_2021_ICCV,boss2021neural,srinivasan2021nerv,physg2020,munkberg2021extracting}.
These assumptions limit existing methods from recovering accurate and rich scene properties, resulting in limited re-rendering results as well, e.g. global illumination effects cannot be modeled. 
%
%
\par 
In parallel, the field of Computer Graphics has extensively researched the problem of photorealistic image synthesis. 
These methods take a well-defined 3D scene and render a realistic image.
Methods have explored different ways of modeling indirect illumination using path tracing. 
Since most path tracing methods are inefficient, precomputed radiance transfer (PRT) was introduced as an efficient approximation of global illumination~\cite {sloan2002precomputed,ritschel2012state,sloan2005local,wang2018,wu2020analytic,lehtinen2007framework} . 
However, these approaches usually do not consider \textit{recovering the PRT solely from images}.
%
%
\par 
In this paper, we combine the learning of precomputed radiance transfer function with inverse rendering, thus combining the best of Computer Vision and Computer Graphics. 
The precomputed radiance transfer function is parameterized as a neural network. 
Thus, it does not require any predefined approximation function, e.g. spherical harmonics. 
As we model the material using a learned PRT, our method does not share common limitations with existing inverse rendering methods -- our method is capable of dealing with complex light paths such as indirect reflections and shadows, and is also not limited to any predefined BRDF model. 
Our method is learned on multi-view observations of a scene under a single unknown light condition. 
In addition to the PRT, it also recovers the scene illumination as an environment map, and the scene geometry defined as a neural signed distance field. 
Thus, our method enables applications such as novel-view synthesis and global relighting using environment maps. 
Existing methods, which enable these applications while taking global illumination into account, rely on light-stage datasets, where the object is captured from multiple views under different light conditions. 
We show that such a setup is not essential. 
In contrast to real light-stage data, we generate synthetic light-stage data of the scene using a high-quality renderer. 
This enables correct disentanglement of the material and illumination properties in the scene, while the real multi-view data allows us to capture photorealistic details and to overcome the common assumptions made by the renderer. 
In summary, our contributions are:
\begin{itemize}
\item{A method for recovering the radiance transfer field from images of objects under an unknown light condition, hence enabling free-viewpoint relighting with realistic global illumination.}
\item{A \textit{neural} precomputed radiance transfer (PRT) field for multi-bounce global illumination computation and neural implicit surface rendering.}
\item{A novel supervision strategy leveraging a differentiable ray-tracer for physically based scene reconstruction, multiple light bounce rendering, and a new synthetic OLAT supervision.}
\end{itemize}
Our qualitative and quantitative results demonstrate a clear step forward in terms of the recovery of scene parameters as well as the synthesis quality of our approach under novel views and lighting conditions when comparing to the previous state of the art.
We will make the code and the new dataset publicly available.
%
%
\section{Related Work}
\label{sec:related}
In the following, we focus on previous work concerning radiance transfer and inverse rendering.
Although our method also recovers the scene geometry using an off-the-shelf implicit geometry reconstruction approach~\cite{wang2021neus}, it is not the main focus of this work and, thus, we do not review related work in this area.
%
%
\paragraph{Precomputed Radiance Transfer.}
%
%
Precomputed Radiance Transfer (PRT) \cite {sloan2002precomputed} is a powerful approach for efficient rendering of global illumination \cite{ritschel2012state}.
Typically, static geometry and reflectance in combination with distant illumination are assumed, which allows to partially precompute light transport for free-viewpoint synthesis and dynamic lighting.
Extensions to \eg dynamic objects \cite{sloan2005local} or near-field illumination \cite{wang2018,wu2020analytic} exist.
The generic formalization of PRT \cite{lehtinen2007framework} enables the incorporation of arbitrarily complex light transport, including multi-bounce light paths. 
While these works improve the runtime of the forward rendering pipeline, they do not consider recovering the PRT solely from a set of real world images of the object.
In contrast, we employ this concept to efficiently decompose illumination and reflectance for view synthesis and relighting, taking into account full global illumination, but apply these concepts in an inverse setting where we aim to recover the PRT from images by means of training a neural PRT network.
%
%
%
\par 
The versatility of PRT has encouraged the exploration of different angular basis functions, such as spherical harmonics~\cite{sloan2002precomputed,kautz2002fast}, Haar wavelets~\cite{ng2003all}, spherical isotropic~\cite{tsai2006all} and anisotropic~\cite{xu2013anisotropic} radial basis functions. 
While the inherent prior of such basis functions can be beneficial for inverse problems, they also limit the range of illumination effects that can be explained by such a basis.
As a remedy, we use the primal directional basis \cite{hao2003interactive} in combination with a neural network, to overcome the limitations of classical basis functions.
We encode the full radiance transfer into a neural field \cite{tewari2021advances,xie2021neural}. 
Recently, Rainer \etal \cite{rainer2022} have explored PRT-inspired neural field-based forward rendering of synthetic scenes -- with full knowledge of all scene parameters, as in most works discussed above. 
In contrast, our framework is concerned with global illumination-aware novel-view synthesis and relighting from multi-view data under one unknown illumination condition.
Further, the transfer from distant illumination to local lighting is traditionally concerned with the \emph{incoming} radiance at a surface point \cite{kautz2005precomputed}, \ie the convolution with reflectance is excluded from precomputation to increase efficiency and to reduce storage requirements.
In contrast, we follow ideas from PRT-based relighting \cite{ng2003all,thul2020precomputed} in that we directly predict \emph{outgoing} radiance. 
%
%
\paragraph{Inverse Rendering and Relighting.}
%
%
Inverse rendering \cite{marschner1998inverse,ramamoorthi2001signal} aims at estimating scene properties such as geometry, lighting, and materials from image observations. 
In this work, we are particularly interested in decoupling lighting using multi-view data, and therefore focus our literature review on corresponding related work in illumination decomposition and relighting.
%
%
%
\par
Controllable illumination in a multi-view light stage \cite{debevec2000acquiring} is conceptually the most straightforward way of obtaining a light-reflectance decomposition in the presence of global illumination, via one-light-at-a-time (OLAT) captures. 
Even though not trivial, novel-view synthesis and relighting boils down to clever interpolation \cite{sun2020light,zhang2021neural,Pandey2021}.
In contrast, input to our method is casually-captured multi-view data under unknown illumination, while embedding \emph{synthetic} OLAT data generation into the training process to aid disentanglement.
%
%
%
\par
Techniques for inverse rendering from multi-view data typically impose strong assumptions on lighting and material, with shading models commonly only considering \emph{direct} illumination \cite{philip2019multi,zhang2021nerfactor,Boss_2021_ICCV,boss2021neural,srinivasan2021nerv,physg2020,munkberg2021extracting}.
Different scene representations have been explored in this context, including meshes \cite{philip2019multi,munkberg2021extracting}, signed distance functions (SDFs) \cite{physg2020}, or neural radiance or reflectance fields \cite{Mildenhall20eccv_nerf,zhang2021nerfactor,Boss_2021_ICCV,boss2021neural,srinivasan2021nerv}.
A common paradigm is the explicit reconstruction of a material representation, \eg an albedo and roughness map, limiting them to recover appearance effects within the range of these predefined representations.
In contrast, our approach seeks to decompose observed color into illumination and a radiance transfer function in a surface-based scene representation, enabling relighting with intricate \emph{indirect} illumination, while reconstructing materials only for supervision.
%
%
\par
Incorporating multiple light bounces into inverse rendering and relighting can be done by using heuristic lighting models \cite{laffont2012rich,lyu2021efficient}, by assuming known illumination \cite{goel2020shape}, or by employing physically-based rendering to approximate irradiance \cite{philip2021free}.
 Chen \etal  \cite{chen2020neural} approximate PRT in neural rendering, given geometry, without physically-based modeling of multiple light bounces.
Thul \etal \cite{thul2020precomputed} utilize PRT in a custom optimization to perform global illumination-aware decomposition of lighting and materials, approximating the required gradients with a single-bounce model.
In contrast, differentiable path tracing \cite{nimier2019mitsuba,li2018differentiable,BangaruMichel2021DiscontinuousAutodiff} can be used to obtain full gradients for global illumination-aware inverse rendering \cite{azinovic2019inverse,nimier2021material}. 
We also leverage the concept of differentiable path tracing \cite{nimier2019mitsuba} during training as a means for achieving disentanglement.
Different from path tracing, our performance at inference time is independent of light transport complexity and by design produces noise-free renderings of multi-bounce illumination. 
%
%
\section{Method} \label{sec:method}
%
%
%
%
%
\begin{figure}
	\begin{center}
		\includegraphics[width=\linewidth]{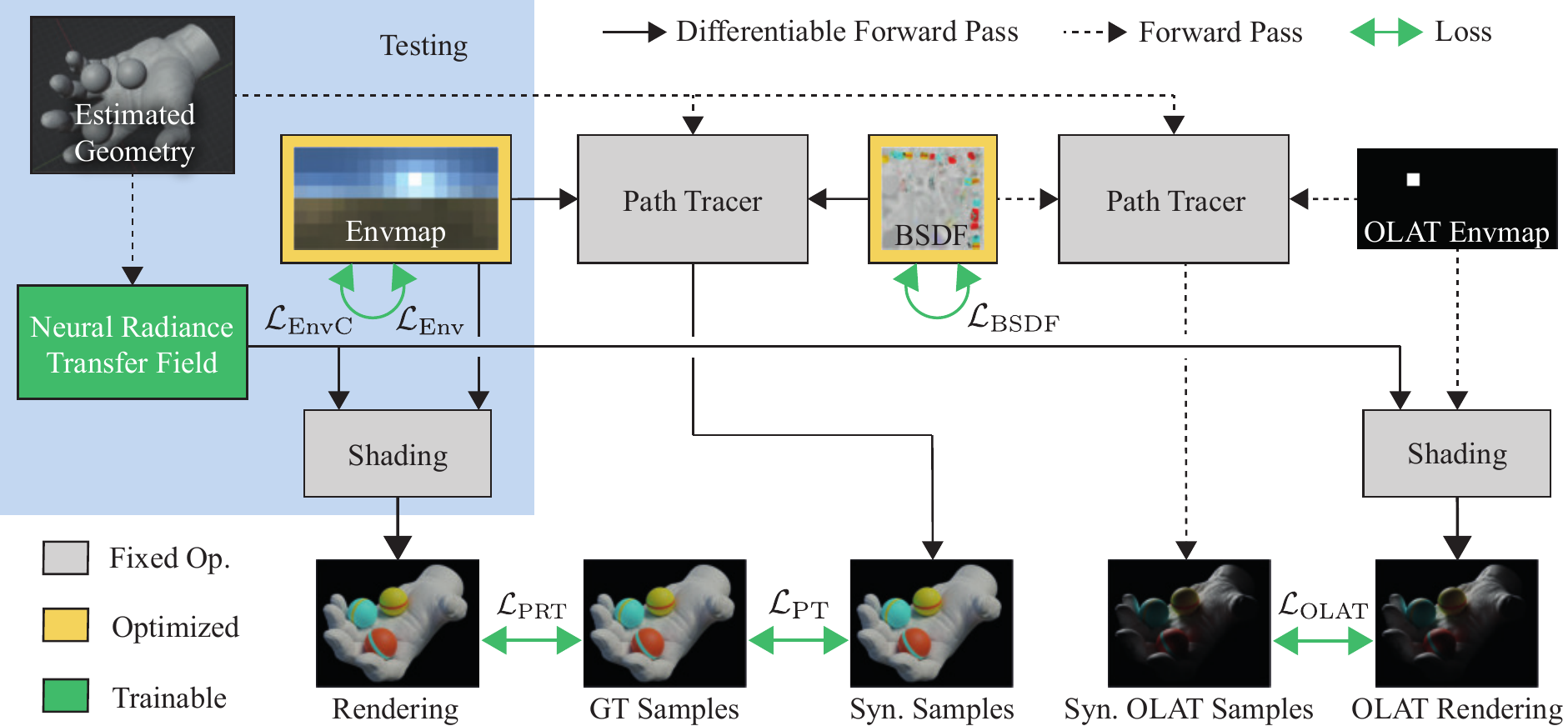}
	\end{center}
	\vspace{-12pt}
	\caption
	{
	Overview of our pipeline. The Shading blocks evaluate Eq.~\ref{eq:prt}. The directional inputs to the radiance transfer field are omitted to avoid clutter. The blue area marks the parts run at test time, when any environment map can be used to light the scene with full global illumination.
	}
	\label{fig:overview}
\end{figure}
%
%
%
%
Our method takes $m \approx 64$ posed multi-view images under an unknown illumination condition as input and allows efficient novel-view synthesis and relighting for the object depicted in these images.
To this end, our approach leverages the concept of precomputed radiance transfer (PRT) to factor multi-view observations into illumination and reflectance.
Thus, at inference, novel illumination conditions in the form of environment maps can be multiplied with our learned reflectance field given a user-defined camera viewpoint. 
%
%
\par 
In more detail, the rendering equation and its equivalent formulation in the PRT framework forms the theoretical foundation of our approach, and we provide a brief overview of it (Sec.~\ref{subsec:render_equ}).
Then, we introduce our neural radiance transfer field (NRTF), a neural network that takes as input a point on the surface and its normal, as well as the incoming and outgoing light directions and predicts the radiance transfer of the scene.
This radiance can then be multiplied with an arbitrary environment map enabling global illumination relighting (Sec.~\ref{subsec:PRT}).
To achieve this, we first estimate scene geometry from the available multi-view observations using implicit surface reconstruction \cite{wang2021neus} (Sec.~\ref{subsec:shape}).
In order to train the NRTF, an approximate disentanglement between the observed material and lighting in the multi-view training images is required.
Our solution for this is to leverage a differentiable path tracer \cite{nimier2019mitsuba}.
It allows the joint optimization of a spatially-varying BSDF and the environment map (Sec.~\ref{subsec:light&material}).
Once the BSDF is obtained, the path tracer can be used to synthesize one-light-at-a-time (OLAT) renderings of the scene (Sec.~\ref{subsec:OLAT}).
The NRTF is trained using a combined loss, consisting of a real image loss that helps to recover photoreal material effects beyond the effects possible with the BSDF model, as well as a synthetic OLAT loss that acts as a prior improving generalization to novel lighting conditions (Sec.~\ref{subsec:trainingNRTF}).
An overview of our pipeline is shown in Fig.~\ref{fig:overview}.
%
%
\subsection{Background} \label{subsec:render_equ}
We are interested in estimating radiance $L$ arriving from scene point $\mathbf{x} \in \mathbb{R}^3$ in direction $\boldsymbol{\omega}_{o} \in \Omega$, where $\Omega$ denotes the space of 3D directions, \ie points on the unit sphere. 
The rendering equation \cite{kajiya1986rendering} describing global light transport can be formulated as
%
%
\begin{equation}
L\left(\mathbf{x}, \boldsymbol{\omega}_{o}\right)=\int_{\Omega_+} L\left(x, \boldsymbol{\omega}_{i}\right) \rho\left(\mathbf{x}, \boldsymbol{\omega}_{i}, \boldsymbol{\omega}_{o}\right) \left(\boldsymbol{\omega}_{i} \cdot \mathbf{n}\right) d \boldsymbol{\omega}_{i} \enspace{,} 
\end{equation}
%
%
where $\Omega_+$ is a hemisphere centered at the surface normal $\mathbf{n}$ of $\mathbf{x}$, $\boldsymbol{\omega}_i$ is an incoming direction, and $\rho$ is the bidirectional scattering distribution function (BSDF) encoding spatially-varying surface material reflectance.
Solving this integral equation including global illumination, \ie multiple light bounces with potentially complex inter-reflections, lends itself to a recursive algorithm like path tracing \cite{kajiya1986rendering}, which stochastically samples light paths to obtain a Monte Carlo estimate of the solution.
While modern differentiable variants of path tracing for inverse rendering \cite{nimier2021material} show promising results, they suffer from high computational costs, especially in the presence of complex light paths.
To gain efficiency, we consider distant but otherwise arbitrary illumination, \ie lighting that can be modeled using an environment map.
Therefore, inspired by PRT, we rewrite the rendering equations as
%
%
\begin{equation}
\label{eq:prt}
L\left(\mathbf{x}, \boldsymbol{\omega}_{o}\right)=\int_{\Omega_+} L_\mathrm{e}\left(\boldsymbol{\omega}_{i}\right) T\left(\mathbf{x}, \mathbf{n}, \boldsymbol{\omega}_{i}, \boldsymbol{\omega}_{o}\right)  d \boldsymbol{\omega}_{i} \enspace{,} 
\end{equation}
%
%
where $L_{e}\left(\boldsymbol{\omega}_{i}\right)$ is the incoming environment light from direction $\boldsymbol{\omega}_{i}$, which is notably independent of $\mathbf{x}$.
The crucial ingredient of this formulation is the collapsed radiance transfer function $T$, which transforms the global distant illumination $L_{e}$ from direction $\boldsymbol{\omega}_{i}$ into local reflected radiance at position $\mathbf{x}$ into direction $\boldsymbol{\omega}_{o}$.
Given an environment map and the scene-specific transfer function $T$, all that is needed to compute global illumination for a pixel is to obtain the primary intersection point $\mathbf{x}$, evaluate $T$ for all environment map texels, multiply with the respective illumination, and sum all contributions.
If $T$ is compact and easy to evaluate, arbitrarily complex global illumination can be efficiently computed on a GPU in a map-reduce fashion.
%
%
\subsection{Neural Radiance Transfer Field (NRTF)} \label{subsec:PRT}
We model the radiance transfer function $T$ using our neural radiance transfer field
%
%
\begin{equation}
\label{eq:nrtf}
    T_\theta
\left(
\mathcal{H}(\mathbf{x}),
\mathbf{n},
\mathcal{F}(\boldsymbol{\omega}_{i}),
\mathcal{F}(\boldsymbol{\omega}_{o})
\right)
=
\mathbf{c}_t \mathrm{,}
\end{equation}
%
%
where $\mathbf{c}_t \in \mathbb{R}^3$ denotes transferred RGB color and $\theta$ indicates the trainable parameters.
We parameterize $T_\theta$ using a multi-layer perceptron (MLP).
Here, we apply a multi-resolution hash encoding $\mathcal{H}(\cdot)$ \cite{mueller2022instant} to the 3D position $\mathbf{x}$, and a spherical harmonics encoding $\mathcal{F}(\cdot)$ \cite{yu2021plenoxels} to light directions $\boldsymbol{\omega}_{i}$ and $\boldsymbol{\omega}_{o}$.
The hash encoding enables faster training and evaluation of our networks.
Details about the network architecture and the encoding strategy can be found in the supplemental document.
When rendering an image from an arbitrary camera view centered at $\mathbf{o}$, we shoot a ray $\mathbf{r}(t)= \mathbf{o} + t \boldsymbol{\omega}_{o}$ through a pixel with 2D coordinate $\mathbf{u}$, and compute the intersection point $\mathbf{x}$ with respect to the scene geometry. 
At $\mathbf{x}$, we now evaluate a discretized version of Eq.~\ref{eq:prt} using our learned $T_\theta$:
With $\hat{\boldsymbol\omega}$ denoting discrete incoming directions, corresponding to the pixels of a discretized environment map $\hat{L}_\mathrm{e}$, we write
%
%
\begin{equation}
\label{eq:prt_discretized}
L_{\theta}(\mathbf{u})
=
L_\theta\left(\mathbf{x}, \boldsymbol{\omega}_{o}\right)
=
\sum_{\hat{\boldsymbol\omega}_i} 
\hat{L}_\mathrm{e}(\hat{\boldsymbol\omega}_i)
\cdot
T_\theta
\left(
\mathcal{H}(\mathbf{x}),
\mathbf{n},
\mathcal{F}(\hat{\boldsymbol\omega}_i),
\mathcal{F}(\boldsymbol{\omega}_{o})
\right).
\end{equation}
%
%
Note that this process is repeated for each pixel of the output image.
It is worth emphasizing again that this formulation can capture multi-bounce lighting effects and complex material reflectance.
Importantly, $\mathbf{x}$, $\boldsymbol{\omega}_{o}$, and $\hat{L}_\mathrm{e}$, \ie the camera and the environment map can be modified at test time, enabling free-viewpoint rendering and scene relighting.
In the following, we explain how we first obtain the scene geometry from the set of multi-view images and then provide details on how the NRTF can be trained without ground truth scene lighting and material.
%
%
\paragraph{Geometry Estimation.} \label{subsec:shape}
In general, our approach is agnostic to the type of surface-based geometry representation.
Recent neural rendering methods~\cite{tewari2021advances} have demonstrated state-of-the-art shape reconstruction results using implicit neural SDF representations. 
We leverage the recently proposed NeuS~\cite{wang2021neus} for computing the SDF geometry of the object.
NeuS takes multi-view images and camera poses as input and reconstructs the geometry, represented as a neural field.
Since rendering an explicit mesh is significantly more efficient than rendering an SDF, we extract a mesh from the implicit surface using Marching Cubes \cite{lorensen1987marching} and use this mesh in our method. We utilize Blender’s ``Smart UV Project'' operator \cite{Blender} to automatically generate the texture map for the mesh extracted from the SDF.
%
%
\subsection{Path Tracing for Initial Light and Material Estimation} \label{subsec:light&material}
As an initial step, we leverage the state-of-art differentiable path tracer Mitsuba~2~\cite{nimier2019mitsuba} to optimize material properties and scene illumination.
We choose a blended BSDF type, where a rough conductor BSDF with roughness $\alpha$ and a diffuse BSDF with a $512\times512$ texture $A$ is combined using a convex combination with weight $w$.
Illumination is represented as a $32\times16$ environment map $\hat{L}_\mathrm{e}$.
Jointly optimizing light and material properties is difficult due to the ambiguities in the image formation process. 
In order to make our optimization stable, we assume the object to have a specular material that does not vary spatially.
However, we use a spatially-varying diffuse texture $A$ for capturing details. 
While these assumptions are often incorrect for many complex scenes, we show that our neural radiance transfer function is capable of reconstructions beyond these assumptions.
Using the reconstructed geometry, it is straightforward to obtain foreground masks for each input view $I_i$, and we define the set of all foreground pixels as $\mathcal{M}_i$.

We jointly optimize ${w,\alpha,A,\hat{L}_\mathrm{e}}$ from the input multi-view images and the precomputed geometry, using the following loss term:
%
%
\begin{align}
    \mathcal{L}({w,\alpha,A,\hat{L}_\mathrm{e}}) = \mathcal{L}_\text{PT} ({w,\alpha,A,\hat{L}_\mathrm{e}}) + \lambda_\text{reg}  \mathcal{L}_\text{reg}({A,\hat{L}_\mathrm{e}}) \,.
\end{align}
%
%
It consists of a data term and a regularizer that is weighted by $\lambda_\text{reg}$. The data term is defined as 
%
%
\begin{align}
    \mathcal{L}_\text{PT} ({w,\alpha,A,\hat{L}_\mathrm{e}}) = \sum_{i=1}^m \sum_{\mathbf{u} \in \mathcal{M}_i} ||{\hat{I}_i (\mathbf{u}) - I_i(\mathbf{u})}||^2 \,.
\end{align}
%
%
Here, $\hat{I}_i$ is the path-traced reconstruction using up to five light bounces from the estimated scene parameters rendered from the $i$th viewpoint, $\mathbf{u}$ denotes 2D pixel coordinates, and $|| \cdot ||$ is the Euclidean norm.
We use two regularizers to better constrain the problem:
%
%
\begin{align}
    \mathcal{L}_\text{reg} (A,\hat{L}_\mathrm{e}) = \sum_{\hat{\boldsymbol\omega}_i}
    | \nabla \hat{L}_\mathrm{e}(\hat{\boldsymbol\omega}_i) | 
    +  \lambda_\text{BDSF}  \sum_{\mathbf{u'} \in \mathcal{M}_\mathrm{tex}} 
    | \nabla {A} (\mathbf{u'}) |
    \end{align}
%
%
where $\nabla(\cdot)$ denotes the image gradient, $\lambda_\text{BDSF}$ is a weighting factor, and $| \cdot |$ denotes the L1 norm.
$\mathbf{u'}$ are 2D $uv$-coordinates in the texture map and $\mathcal{M}_\mathrm{tex}$ is the set of texels that is covered by the unwrapped geometry.
The first term, $\mathcal{L}_\text{Env}$, is a regularizer on the environment map reconstruction, while the second term, $\mathcal{L}_\text{BSDF}$, regularizes the texture reconstruction.
Both encourage image gradient sparsity. 
We refer to the supplemental document for more details.
%
%
\subsection{Training the Neural Radiance Transfer Field} \label{subsec:trainingNRTF}
\paragraph{OLAT Synthesis.} \label{subsec:OLAT}
Our goal is to train the neural transfer field for the input scene. 
If we train the neural network only with the input illumination condition, the network can easily overfit, thus, not being able to disentangle illumination and material.
Traditionally, learning-based methods, which disentangle material and illumination properties, rely on light-stage capture setups~\cite{debevec2000acquiring,sun2020light,zhang2021neural,Pandey2021}.
In contrast to these approaches, we only rely on a single illumination condition. 
We show that it is possible to train for disentanglement even in this more challenging setup, by simulating a virtual light stage. 
Using the reconstruction obtained with the differentiable path tracer, we render synthetic images of the scene under novel one-light-at-a-time (OLAT) illumination conditions. 
Here, only one pixel on the environment map is active at a time. 
We sample OLAT images for training and novel camera views from every incoming light direction. 
We use Blender \cite{Blender} to render the OLAT images with the reconstructed geometry and material as input.
In total, we synthesize $N_\mathrm{c}*N_\mathrm{e}$ OLAT images as extra supervision, where  $N_\mathrm{c}$ is the number of sampled camera views and $N_\mathrm{e}$ is the number of texels in the environment map. 
Note that the OLAT representation forms a complete basis for illumination conditions, i.e., any environment map can be computed as a linear combination of OLAT environment maps.
Using these OLATs for our network supervision enables generalization to unseen illumination conditions and camera views. 
%
%
\paragraph{NRTF Training.}
We train our NRTF in two stages.
First, we train on the OLAT dataset using the following loss:
%
%
\begin{equation}
\label{eq:olat_loss}
\mathcal{L}_\text{OLAT}(\theta)=\sum_{i=1}^{N_\mathrm{c}} \sum_{\mathbf{u}\in \mathcal{M}_\mathrm{i} } 
\left| \left| 
\frac{L_{\theta,i} (\mathbf{u}) -O_{i} (\mathbf{u})}{\operatorname{sg}\left(L_{\theta,i} (\mathbf{u})  \right)+\epsilon}
\right|\right|^{2} \,,
\end{equation}
%
%
Here, $O_i$ is the $i$th OLAT image from Blender and $L_{\theta,i}$ is the corresponding estimate from our NRTF using Eq.~\ref{eq:prt_discretized}.
Stop gradient is denoted by $\operatorname{sg}(\cdot)$.
We set $\epsilon=1e-3$ to avoid division by zero and optimize for the network parameters $\theta$.
As shown in Noise2Noise~\cite{lehtinen2018noise2noise}, this loss works better for high-dymanic range images in the presence of path-tracing noise.
Training on the OLAT images enables relighting and novel-view synthesis using the learned transfer function. 
\par 
However, the method so far is heavily influenced by the lighting-reflectance ambiguity, and by the assumption of a global specularity parameter.
Thus, in a second step, we further finetune the network on the input multi-view images. 
Here, we sample images from the real images as well as the synthetic OLAT images in a minibatch for training. 
The loss for this stage is defined as
%
%
\begin{align}
    \mathcal{L}(\theta, \tilde{L}_\mathrm{e}) = 
    \mathcal{L}_\text{OLAT}(\theta, \tilde{L}_\mathrm{e}) 
    + \lambda_\text{PRT} \mathcal{L}_\text{PRT} (\theta, \tilde{L}_\mathrm{e})
    + \lambda_\text{EnvC}  \mathcal{L}_\text{EnvC} (\tilde{L}_\mathrm{e}) \,.
\end{align}
%
%
$\mathcal{L}_\text{OLAT}$ is used for the OLAT images in the batch.
It is defined as in Eq.~\ref{eq:olat_loss}, however, we also finetune the environment map $\tilde{L}_\mathrm{e}$ in this stage using the previously obtained environment map $\hat{L}_\mathrm{e}$ for initialization.
We further use a masked L2 loss for real images as:
%
%
\begin{align}
    \mathcal{L}_\text{PRT} (\theta, \tilde{L}_\mathrm{e}) = \sum_{i=1}^m \sum_{\mathbf{u} \in \mathcal{M}_i} 
    ||{L_{\theta,i} (\mathbf{u}) - I_i(\mathbf{u})}||^2 \,,
\end{align}
%
%
Training on real images allows us to update the environment map. 
We add a regularizer, which penalizes the output to be too far from the initial environment map.
%
%
\begin{align}
    \mathcal{L}_\text{EnvC} (\tilde{L}_\mathrm{e}) = \sum_{\hat{\boldsymbol\omega}_i} ||\tilde{L}_\mathrm{e} (\hat{\boldsymbol\omega}_i)) - \hat{L}_\mathrm{e} (\hat{\boldsymbol\omega}_i)||^2 \,,
\end{align}
%
%
where $\hat{L}_e (\hat{\boldsymbol\omega}_i)$ denotes the initial environment map estimate. 
%
%
%
\section{Results} \label{sec:results}
%
%
Next, we report results of the experiments we conducted to evaluate our method.
We construct five synthetic scenes to showcase global illumination effects and further utilize four real scenes from the DTU dataset \cite{jensen2014large}.
For each scene, we take 32-64 input views with a resolution of $800 \times 600$ pixels.
During training, all our environment maps have a resolution of $32 \times 16$ pixels in latlong format, but this resolution can be different at test time due to our continuous neural-field formulation.
On a single Quadro RTX 8000 GPU, training takes half an hour for initial light and material estimation, eight hours for OLAT training, and an additional 16 hours for the final joint optimization to reach highest-quality results.
%
%
Factorizing lighting and reflectance is fundamentally ambiguous \cite{land1971lightness} and cannot be resolved from image observations in general \cite{ramamoorthi2004signal,lombardi2015reflectance}, especially when allowing for spatially-varying materials \cite{ramamoorthi2001signal}. 
To aid meaningful comparisons nevertheless, we follow  the procedure of Zhang \etal \cite{zhang2021nerfactor} and other works for all qualitative and quantitative results on synthetic scenes: 
We compute the mean RGB value of the ground truth environment map and normalize our estimated lighting by its inverse.

%
%
\subsection{Qualitative Results} \label{sec:qualitativeresults}
%
%
%
\begin{figure}[!t]
	\begin{center}
		\includegraphics[width=\linewidth]{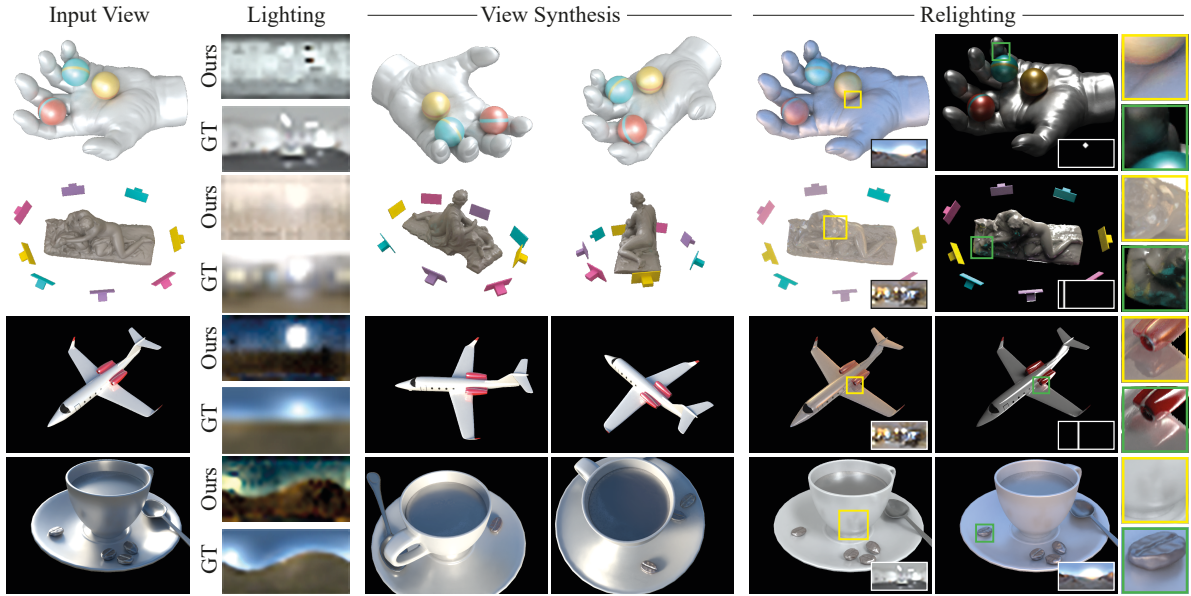}
	\end{center}
	\vspace{-12pt}
	\caption
	{
	Qualitative results on synthetic data.
	First column: Example input view for training.
	Second column: Our estimated environment map and the corresponding ground truth.
	Third and fourth column: Novel view synthesis results of our approach. Note that our method achieves sharp and accurate novel views that are almost indistinguishable from the input views in terms of quality.
	Last two columns: Relighting results of our method using novel environment maps (see insets). Also for novel lighting conditions our approach achieves convincing results with sharp specular reflections and secondary light bounce effects, e.g. indirect reflections on the wing of the airplane.
	}
	\label{fig:qualitative}
\end{figure}
%
%
%
In Fig.~\ref{fig:qualitative}, we demonstrate qualitative results of our method.
Next to a representative input view (first column), we show the estimated and ground truth lighting (second column), followed by two exemplary novel views created with our method (third and fourth column).
Finally, we show relighting results (remaining two columns) using different environment maps (insets).
We see that our method produces high-quality relightable novel views, while successfully incorporating global-illumination effects like higher-order specular reflections and subtle color bleeding (see also Fig.~\ref{fig:teaser}).
In our supplemental video, we further show that our method is also temporally stable when smoothly changing the camera view or rotating an environment map.
\paragraph{Real Data.}
%
%
%
\begin{figure} [!t]
	\begin{center}
		\includegraphics[width=\linewidth]{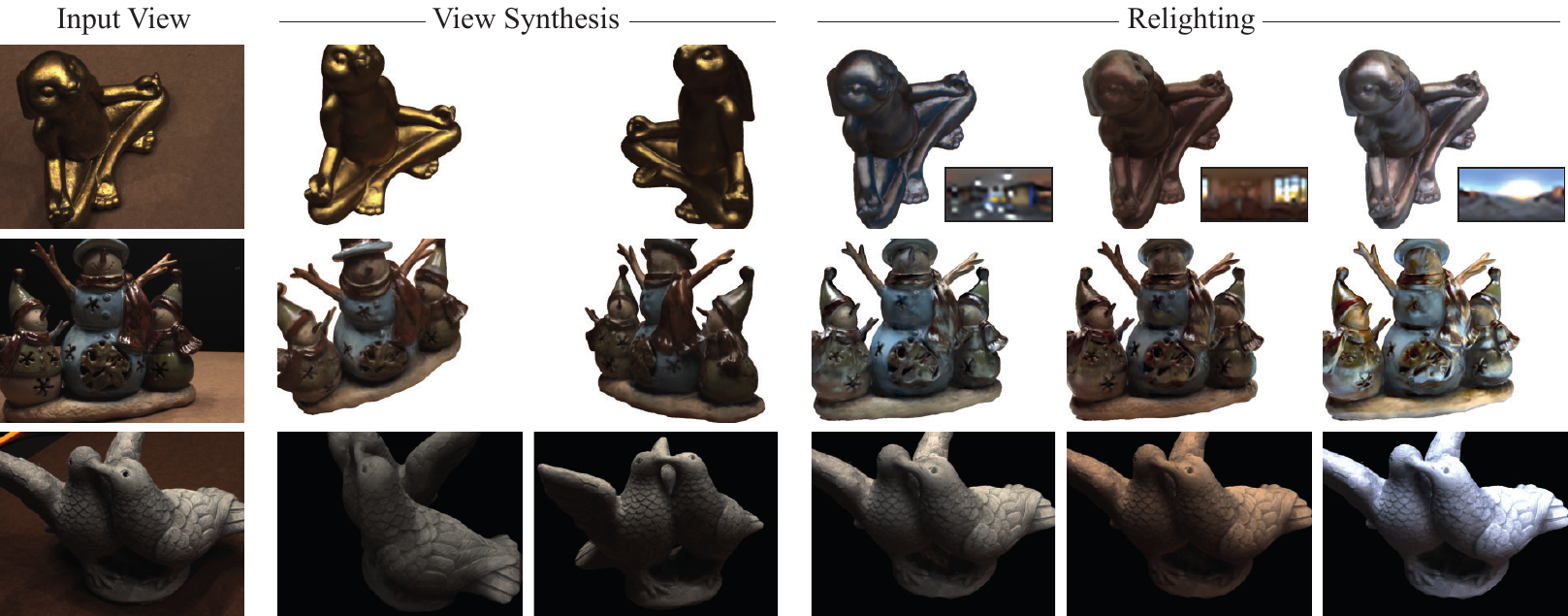}
	\end{center}
	\vspace{-12pt}
	\caption
	{
	Qualitative results on real data~\cite{jensen2014large}.
	First column: Example input views used for training our method.
	Second and third column: Novel view results. Note that even for real data our method achieves realistic novel view renderings.
	Last three columns: Relighting results using the environment maps depicted in the insets.
	Also here, note that our method can achieve plausible relighting effects.
	}
	\label{fig:realdata}
\end{figure}
%
%
%
In Fig.~\ref{fig:realdata}, we show results of our method on real scenes of the DTU dataset \cite{jensen2014large}.
We can successfully synthesize high-quality novel views and plausible relighting.
This shows that our method is robust to such real world captures, which are very challenging due to the lack of very precise camera calibration and foreground segmentation, camera noise, and other effects that are typically not present in synthetic datasets.
%
%
%
%
\begin{figure} [!t]
	\begin{center}
		\includegraphics[width=.85\linewidth]{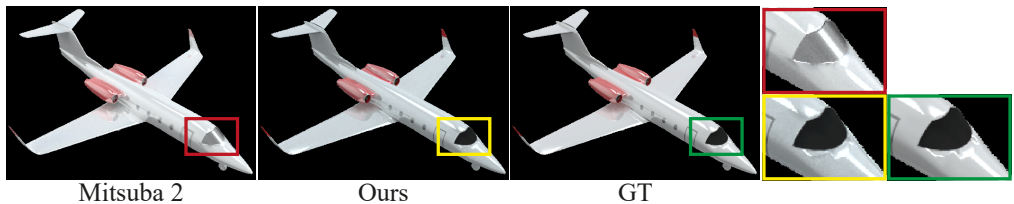}
	\end{center}
	\caption
	{
    Supervision with real input images lets our NRTF (center) learn appearance effects beyond the material model used during initialization with Mitsuba 2 (left), producing images close to the ground truth (right). Notice that our approach captures spatially-varying material roughness, see close-ups. Images show a relit novel view.
	}
	\label{fig:varyingBSDF}
\end{figure}
%
%
%
\paragraph{Beyond the Model Assumptions.}
In Fig.~\ref{fig:varyingBSDF}, we show that our approach can learn spatially-varying material effects beyond the ones that can be explained by the light and material models of the differentiable path tracer~\cite{nimier2019mitsuba}. 
This is due to the real image loss, which lets the network learn appearance effects from real observations.
%
%
\subsection{Comparisons} \label{sec:quantitative}
%
%
%
\begin{table}[!t]
	\begin{center}
    	\label{tab:comparison}	
    	\caption
    	{
    		Numerical comparisons for novel-view synthesis and relighting. We compare to the recent state of the art Mitsuba2 \cite{nimier2019mitsuba}, PhySG \cite{physg2020},  Neural-PIL \cite{boss2021neural}, NeRFactor\cite{zhang2021nerfactor} and RNR \cite{chen2020neural} in terms of image-based metrics, i.e. PSNR and SSIM, and perceptual metrics, i.e. LPIPS.
    		For both tasks, novel view synthesis and relighting, we achieve the best performance.
    	}
		\begin{tabular}{|c|c|c|c|c|c|c|}
			\hline
			 & \multicolumn{3}{|c|}{Novel View Synthesis} & \multicolumn{3}{|c|}{Novel View Synthesis \& Relighting} \\
			\hline
			\textbf{Method}  & \textbf{PSNR} $\uparrow$ & \textbf{SSIM} $\uparrow$ &   \textbf{LPIPS} $\downarrow$  & \textbf{PSNR} $\uparrow$ & \textbf{SSIM} $\uparrow$ &   \textbf{LPIPS} $\downarrow$  \\
			\hline
			Mitsuba2 \cite{nimier2019mitsuba} &23.50 &0.7567 &0.0763 & 21.69&0.5722 &0.0812 \\
			PhySG \cite{physg2020} &20.52   &0.8563 &0.2577 & 17.30 &0.6252  &0.2736 \\
			Neural-PIL \cite{boss2021neural} & 17.07 &0.5563 & 0.1159 &14.76 &0.4895 &0.1328 \\
			NeRFactor \cite{zhang2021nerfactor} &21.97  &0.6394 &0.1691  &15.83 &0.6470 &0.2033 \\
            RNR \cite{chen2020neural} &22.54 &0.8122 &0.0960   & 18.06 & 0.7009 &0.1081 \\
			\textbf{Ours} & \textbf{28.73} & \textbf{0.9151}   & \textbf{0.0454}  & \textbf{23.06}  & \textbf{0.8247}  & \textbf{0.0692}  \\
			\hline
		\end{tabular}
	\end{center}
\end{table}
%
%

%
%
%
\begin{figure}[!ht]
	\begin{center}
		\includegraphics[width=\linewidth]{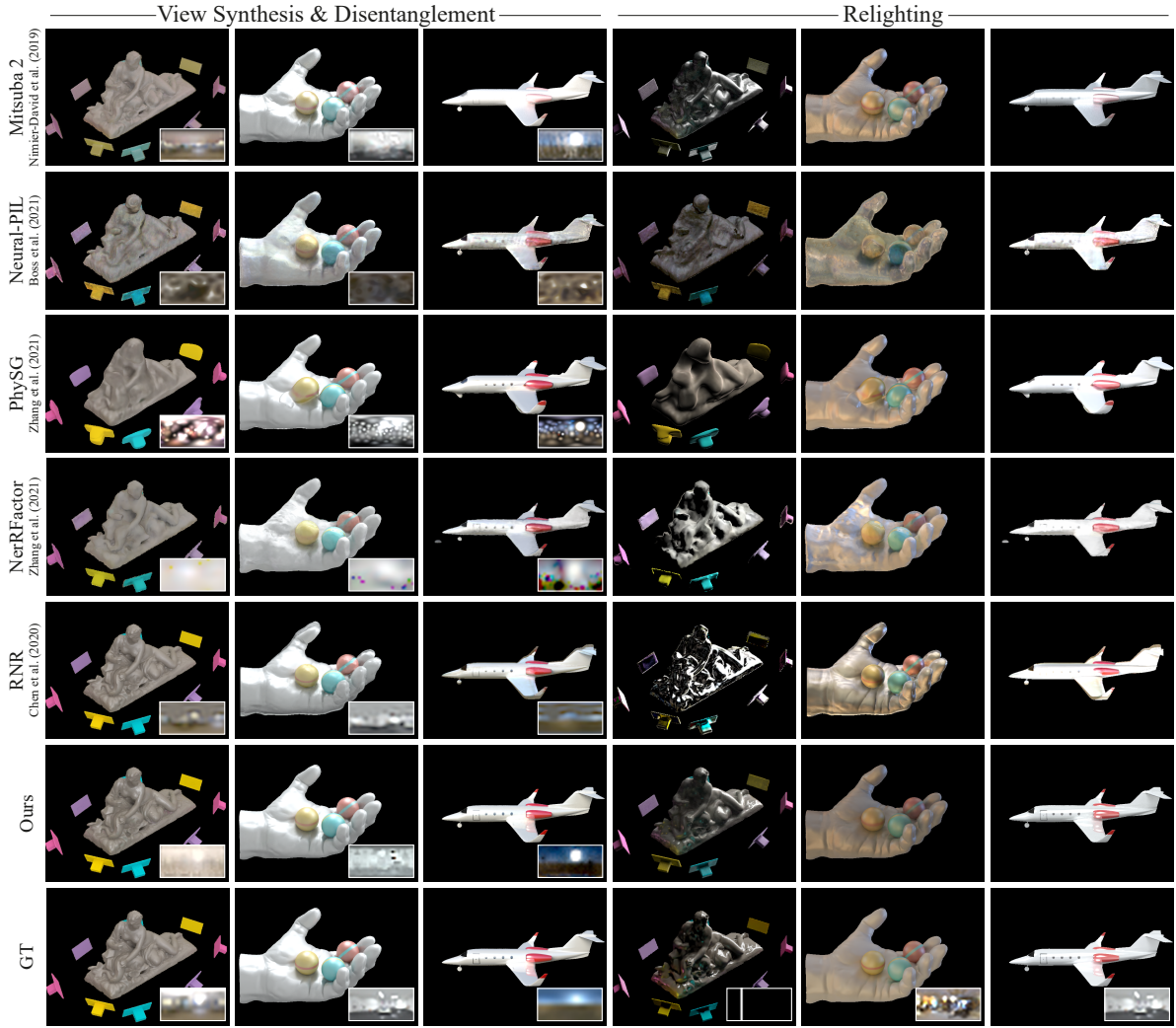}
	\end{center}
	\vspace{-12pt}
	\caption
	{
	Comparisons to related works~\cite{nimier2019mitsuba,physg2020,boss2021neural,chen2020neural,zhang2021nerfactor} for novel-view synthesis and disentanglement of lighting (left three columns), and relighting (right three columns).
	Note that for both tasks we achieve the best results in terms of rendering quality.
	It is also worth noting that we are the only method, which can accurately reproduce the indirect illumination effects such as the self-reflection on the wing of the airplane. 
	}
	\label{fig:comparison}
\end{figure}
%
%
%
We compare our approach against several alternatives on the task of novel-view synthesis with relighting:
On the one hand, we analyze the capabilities of stand-alone differentiable path-tracing using Mitsuba2 \cite{nimier2019mitsuba}, which can be used to perform inverse rendering in the presence of global illumination.
On the other hand, we consider three recent neural field-based inverse-rendering approaches PhySG \cite{physg2020}, Neural-PIL~\cite{boss2021neural}, and NeRFactor~\cite{zhang2021nerfactor}, all employing only a direct illumination model.
We omit a comparison to NeRD \cite{Boss_2021_ICCV} as Neural-PIL can be considered as the follow-up.
We also compare with RNR \cite{chen2020neural}, providing it with the same geometry as ours.
Further, we provide more results on NeRFactor~\cite{zhang2021nerfactor} dataset in the supplemental document.
\par 
A qualitative comparison is shown in Fig.~\ref{fig:comparison}.
Despite the fact that our method provides the sharpest and most realistic results, it is worth noting that our method is the only one that can recover accurate indirect lighting effects, e.g. the self-reflection on the wing of the airplane and the color spill of the squares onto the statue.
This is further confirmed by the quantitative analysis in Table~\ref{tab:comparison}.
We compute image errors for four scenes, each with five views and three lighting conditions according to three metrics on the tasks of novel-view synthesis and novel-view synthesis with relighting.
In particular, we evaluate the Peak Signal-to-noise Ratio (PSNR), the Structural Similarity Index Measure (SSIM)~\cite{1284395}, and the learned perceptual image patch similarity (LPIPS)~\cite{8578166}.
PSNR and SSIM are purely image-based metrics and, thus, sometimes do not reflect the \textit{perceived} image quality. 
For this reason, we also provide the perceptual LPIPS metric.
We observe that our approach again delivers the highest-quality results for both tasks across all metrics.
%
%
\subsection{Ablation \& Extension}
%
%
%
\begin{table}[!t]
	\begin{center}
    	\label{tab:ablation}
    	\caption
    	{
    		Ablations and extensions. 
    		First, we evaluate the effect of the proposed synthetic OLAT loss.
    		One can clearly see that without the OLAT loss the performance of our method drastically drops for the relighting task.
    		This can be explained by the fact that the OLAT loss acts as a regularizer and prevents overfitting to the single environment map that is recovered during training. 
    		Moreover, we evaluate how our method performs when the object was observed under \textit{multiple} lighting conditions.
    		Interestingly, with this additional input, our method can achieve even better results, especially for the relighting task.
    	}
		\begin{tabular}{|c|c|c|c|c|c|c|}
			\hline
			 & \multicolumn{3}{|c|}{Novel View Synthesis} & \multicolumn{3}{|c|}{Novel View Synthesis \& Relighting} \\
			\hline
			\textbf{Method}  & \textbf{PSNR} $\uparrow$ & \textbf{SSIM} $\uparrow$ &   \textbf{LPIPS} $\downarrow$  & \textbf{PSNR} $\uparrow$ & \textbf{SSIM} $\uparrow$ &   \textbf{LPIPS} $\downarrow$  \\
			\hline
		    w/o OLAT Loss &31.35 &0.9668 &0.063 &10.03 &0.4547 &0.4487  \\
			\textbf{Full} &29.56 &0.9418 &0.066 &24.76 &0.8665 &0.069  \\
			Multiple Envmaps &30.62 &0.9166 &0.043 &26.67 &0.9071 &0.047  \\
			\hline
		\end{tabular}
	\end{center}
\end{table}
%
%

%
Here, we study ablations and extensions in order to gain further insights into our system.
All results are compiled into Table~\ref{tab:ablation}, where the evaluation protocol is the same as in Sec.~\ref{sec:quantitative}.
First, we consider omitting the OLAT loss (Sec.~\ref{subsec:OLAT}).
We observe that result quality reduces significantly for the relighting task compared to our full method.
This is due to the poor disentanglement of lighting and reflectance and the fact that the network can overfit to the lighting condition in the training data, which also explains why the novel view synthesis without the OLAT loss is slightly more accurate than our method.
\par 
Second, we investigate the behavior of our approach when input views are captured under \emph{multiple} unknown illumination conditions.
In this experiment, we use three different environment maps.
When reconstructing geometry (Sec.~\ref{subsec:shape}), we select only a subset of multi-view images with the same illumination condition, while during initial light and material estimation (Sec.~\ref{subsec:light&material}), we optimize for three individual environment maps.
Not surprisingly, we observe that this extended setup increases result quality even more compared to our full single-lighting approach.
Yet, it has a significantly less pronounced effect compared to the omission of the OLAT training stage, indicating that our pipeline achieves a solid disentanglement for the single-illumination condition.

%
%
\section{Limitations and Future Work} \label{sec:limitations}
Although our method improves the state of the art in terms of image quality and global illumination handling, it still has some limitations, which open up future work in this direction.
In particular, our method relies on an accurate geometry estimate of the scene and we are not jointly optimizing the scene geometry along with the material and lighting of the scene.
Future work could involve a joint reasoning of all these aspects in a differentiable manner such that optimizing all scene aspects jointly can be achieved.
Further, our relighting results are only correct up to a global scale due to the inherent ambiguity between scene illumination and the object material.
Here, future research could explore a minimal setup required to disentangle such ambiguities, e.g. it may be that a single measurement on the surface can resolve the ambiguity.
Last, our method takes several seconds per frame. 
Ideally, it would run at real time enabling interactive scene relighting with global illumination. 
Thus, exploring more efficient scene representations could be an interesting research branch for the future.
%
%
\section{Conclusion} \label{sec:conclusion}
We presented neural radiance transfer fields, which enable global illumination scene relighting and view synthesis given multi-view images of the object.
At the technical core, our method implements the concept of precomputed radiance transfer that disentangles illumination from appearance.
To this end, we propose a neural radiance transfer field represented as an MLP and show how at train time differentiable path tracing and a dedicated OLAT loss can be used to let the network accurately learn such a disentanglement.
Once trained, our rendering formulation allows novel-view synthesis and relighting, which is aware of global-illumination effects.
Our results demonstrate a clear improvement over the current state of the art while future work could involve further improving the runtime and a joint reasoning of geometry, material, and scene lighting.

\noindent\textbf{Acknowledgements}. We would like to thank Xiuming Zhang for his help with the NeRFactor comparisons. Authors from MPII were supported by the ERC Consolidator Grant 4DRepLy (770784).


%
%
\bibliographystyle{splncs04}
\bibliography{egbib}
\end{document}



\pagestyle{headings}
\mainmatter
\def\ECCVSubNumber{}  

\title{Neural Radiance Transfer Fields for Relightable Novel-view Synthesis with Global Illumination\\
-- Supplemental Document --} 

\titlerunning{Neural Radiance Transfer Fields} 

\author{Linjie Lyu$^1$, Ayush Tewari$^2$, Thomas Leimk\"uhler$^1$, Marc Habermann$^1$, and Christian Theobalt$^1$}
\authorrunning{Lyu et al.}

\institute{$^1$Max Planck Institute for Informatics, Saarland Informatics Campus \\ $^2$MIT}

\maketitle
%
%
%
%
\begin{figure}
	\begin{center}
		\includegraphics[width=\linewidth]{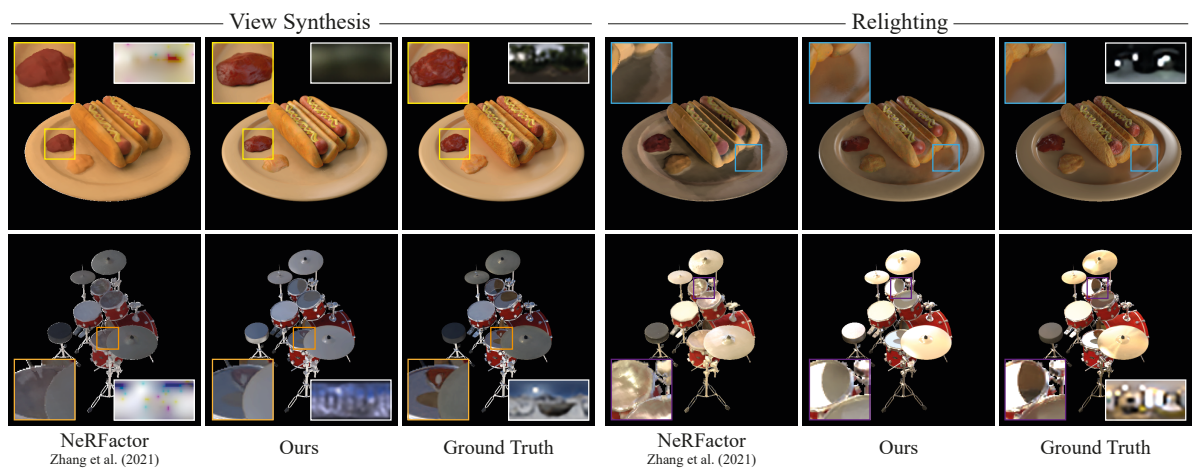}
	\end{center}
	\vspace{-12pt}
	\caption
	{
	Comparison of our approach to NeRFactor \cite{zhang2021nerfactor} on the tasks of novel-view synthesis and relighting. For view synthesis, white insets show estimated (first two columns) and ground truth (third column) environment maps. For relighting, the white inset shows the environment map used to illuminate the scene.
	}
	\label{fig:realdata}
\end{figure}
%
%
%
In the following, we compare our approach to NeRFactor~\cite{zhang2021nerfactor} on their own dataset (Sec.~\ref{sec:comparison}).
Further, we provide more implementation details concerning the network architecture, training, and the regularizers we use (Sec.~\ref{sec:implementation}).
%
%
\section{Comparison with NeRFactor} \label{sec:comparison}
In addition to the comparison in the main document,we provide a comparison with NeRFactor~\cite{zhang2021nerfactor} on their own dataset, i.e. the Drums and Hotdog scenes, in Fig.~\ref{fig:realdata} and in Tab.~\ref{tab:comparison}.
For generating the results for NeRFactor, we used trained models provided by the original authors.
For view synthesis, we randomly sampled 8 views from the 100 test views and computed the average for all metrics.
For relighting, we use the same views and chose the environment map number 3 of their test set.
Note that, visually, our approach synthesizes more detailed images and is able to recover secondary light bounce effects such as the reflection on the drums, which NeRFactor cannot reproduce at all.
This observation is consistent for both the view synthesis and the relighting task.
Moreover, we also found that our method outperforms NeRFactor quantitatively on their own dataset for all metrics, which further confirms the improvement of our method compared to the previous state-of-the-art.
\begin{table}[!t]
	\begin{center}
    	\label{tab:comparison}	
    	\caption
    	{
    		Numerical evaluation for novel-view synthesis and relighting. We compare to the recent state of the art NeRFactor \cite{zhang2021nerfactor} in terms of image-based metrics, i.e. PSNR and SSIM, and perceptual metrics, i.e. LPIPS.
    		PSNR and SSIM were determined on foreground pixels only.
    		For both tasks, novel-view synthesis and relighting, we achieve the best performance.
    	}
		\begin{tabular}{|c|c|c|c|c|c|c|}
			\hline
			 & \multicolumn{3}{|c|}{Novel View Synthesis} & \multicolumn{3}{|c|}{Novel View Synthesis \& Relighting} \\
			\hline
			\textbf{Method}  & \textbf{PSNR} $\uparrow$ & \textbf{SSIM} $\uparrow$ &   \textbf{LPIPS} $\downarrow$  & \textbf{PSNR} $\uparrow$ & \textbf{SSIM} $\uparrow$ &   \textbf{LPIPS} $\downarrow$  \\
			\hline
			NeRFactor \cite{zhang2021nerfactor} &18.67 &0.684 &0.13315 &16.48 & 0.6439 &0.0965 \\
			\textbf{Ours} & \textbf{20.82} & \textbf{0.7577}   & \textbf{0.0659}  & \textbf{20.18}  & \textbf{0.7556}  & \textbf{0.0786}  \\
			\hline
		\end{tabular}
	\end{center}
\end{table}
%
%
\section{Implementation Details} \label{sec:implementation}
Next, we provide more details on the network architecture (Sec.~\ref{sec:net}), the training procedure (Sec.~\ref{sec:train}), and our regularizers (Sec.~\ref{sec:regularizers}).
%
%
\subsection{Network Architecture} \label{sec:net}
%
%
%
\begin{figure}
	\begin{center}
		\includegraphics[width=.8\linewidth]{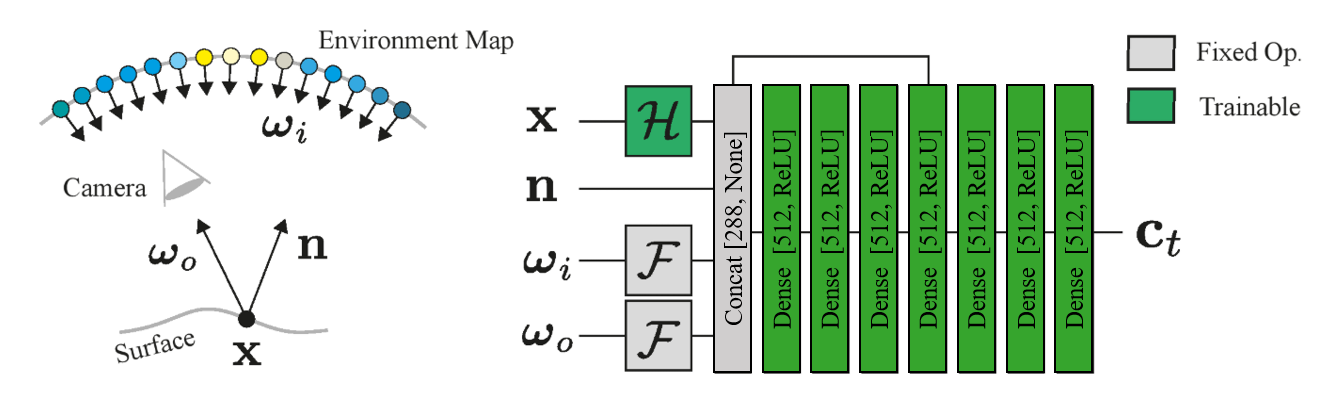}
	\end{center}
	\vspace{-12pt}
	\caption
	{
	The geometry of our setup (left) and our network architecture (right). Expressions in brackets denote [output features, activation function].
	To shade a pixel, we iterate over all incoming directions $\boldsymbol{\omega}_i$. Note that the transferred radiance $\mathbf{c}_t$ takes multiple light bounces into account.
	For details on $\mathcal{H}$ and $\mathcal{F}$ refer to the main text.
	}
	\label{fig:network}
\end{figure}
%
%
%
We model the radiance transfer function $T$ using our neural radiance transfer field
%
%
\begin{equation}
\label{eq:nrtf}
    T_\theta
\left(
\mathcal{H}(\mathbf{x}),
\mathbf{n},
\mathcal{F}(\boldsymbol{\omega}_{i}),
\mathcal{F}(\boldsymbol{\omega}_{o})
\right)
=
\mathbf{c}_t \mathrm{,}
\end{equation}
%
%
which is represented by a multi-layer perceptron (MLP).
First, we encode position $\mathbf{x}$ using a multi-resolution hash function $\mathcal{H}$ as proposed by M\"uller \etal \cite{mueller2022instant}, with parameters $L=16$, $F=16$, $N_\mathrm{min}=16$, and $N_\mathrm{max}=256$. We only create  $1:1$ hash mapping for the voxels near the mesh surface,thus avoiding  a hash collision.
The directions $\omega_i$ and $\omega_o$ are positionally encoded using spherical harmonics features $\mathcal{F}$ \cite{yu2021plenoxels} with $n=4$ frequencies, resulting in 16 features each.
The surface normal $\mathbf{n}$ is fed directly.
To parameterize $T_\theta$, we use an 8-layer MLP, where each hidden layer has 512 features and a ReLU activation function. We additionally use a skip connection between the input layer and the fourth layer.
$\theta$ denotes the trainable parameters of the network and the hash encoding.
A visualization is shown in Fig.~\ref{fig:network}.

%
%
\subsection{Training} \label{sec:train}
Our approach is implemented in Pytorch. 
For training, we use the Adam optimizer~\cite{Kingma2015AdamAM}. 
For optimizing the environment map, the albedo map, and the material coefficients, we optimize for 3000 iterations for each scene with a learning rate of 1 for the environment map, 0.1 for the albedo map, and 0.001 for the material coefficients.
For training our NRTF, during pre-training on the OLAT dataset only, we apply 150k iterations for each scene with a learning rate of 5e-4 and a batch size of 20 OLAT images.
For joint optimization, we apply 100k iterations for each scene with a learning rate of 1e-4. Each batch contains five OLAT images and one real image.
%
%
\subsection{Regularizers} \label{sec:regularizers}
For the image-based regularizers, we approximate image gradients with backward differencing. 
Further, the weights of the inidividual regularizers are
$\lambda_\mathrm{reg} = 0.001$,
$\lambda_\mathrm{BSDF} = 0.1$,
$\lambda_\mathrm{OLAT} = 0.1$,
$\lambda_\mathrm{PRT} = 1$,
$\lambda_\mathrm{EnvC} = 0.001$.
%
%
\bibliographystyle{splncs04}
\bibliography{egbib}